# ArPoMeme: An Annotated Arabic Multimodal Dataset for Political Ideology and Polarization

**Wajdi Zaghouani**[1], **Kais Attia**[2], **Md. Rafiul Biswas**[3], **Fadhl Eryani**[4]
[1]Northwestern University in Qatar, [2] Independent Researcher, Qatar
[3]Hamad Bin Khalifa University, Qatar, [4]University of Tübingen, Germany

**Abstract**

Memes have become a prominent medium of political communication in the Arab world, reflecting how humor, imagery, and text interact to express ideological and cultural positions. Despite the centrality of memes to online political discourse, there is a lack of systematically curated resources for analyzing their multimodal and ideological dimensions in Arabic. This paper presents *ArPoMeme*, a large-scale dataset of approximately 7,300 Arabic political memes categorized by ideological orientation, including Leftist, Islamist, Pan-Arabist, and Satirical perspectives. The dataset captures the diversity of Arabic meme ecosystems by grounding classification in the self-identification of public Facebook pages and groups that produce and disseminate these memes. To ensure both scale and accuracy, we designed a semi-automated data collection pipeline combining Playwright-based Facebook scraping with Google Drive synchronization, followed by text extraction using the Qwen2.5-VL-7B vision–language model. The extracted text was manually verified and annotated for three polarization dimensions: *Us vs. Them framing*, *Hostility toward out-groups*, and *Calls to action*. Annotation was conducted through a custom Streamlit-based interface supporting distributed labeling, real-time tracking, and version control. The resulting dataset links visual content, textual messages, and ideological orientation, enabling fine-grained analysis of political antagonism, mobilization, and humor. Quantitative analysis of the annotated corpus reveals strong asymmetries in antagonistic framing across ideological groups, with Islamist and satirical memes exhibiting the highest levels of hostility and mobilization cues. The dataset and the annotation tool offers a reproducible and publicly available resource for studying Arabic political discourse, multimodal ideology detection, and polarization dynamics.

## 1. Introduction

The rapid growth of digital communication has transformed memes into a central medium of cultural and political expression. Originally circulated primarily for humor, Internet memes have increasingly become politicized, serving as vehicles for commentary, criticism, and ideological positioning (Johann and Bülow, 2019; SHIFMAN, 2014). Drawing from traditions of political cartoons and satire, political memes rely on humor, parody, and juxtaposition to critique authority, reframe narratives, and mobilize audiences (Ali and Mohammed, 2023; Mihăilescu, 2024).

Political memes also serve as a distinctly democratic mode of participation for ordinary citizens. Because sharing or creating memes requires minimal resources, users often describe the practice as an easy, accessible means of engaging in political debate that supplements more conventional forms of online expression and activism (Leiser, 2022). In this sense, memes enable individuals to articulate opinions, contest authority, and connect with like-minded communities in ways that lower the barriers to political engagement. At the same time, the influence of memes extends beyond grassroots circulation: they have been strategically adopted by political elites including parties and candidates who recognize their potential to frame issues and reach audiences in compelling, highly shareable formats (Romo, 2023-04-25). This dual character, at once bottom-up and top-down, positions memes as a powerful driver of contemporary political discourse (McLoughlin and Southern, 2021).

In the Arab world, memes have emerged as a dynamic arena for online political expression. They both mirror and shape narratives surrounding political actors, ideological movements, and contentious events. Despite growing scholarly attention to Arabic memes in the domains of propaganda, disinformation, and hate speech (**?**), no existing dataset systematically captures the ideological diversity of memes circulating within Arabic political discourse. This gap limits our ability to examine how memes mediate ideological struggles across competing currents such as leftist, Islamist, and pan-Arabist movements.

In addition to these clearly defined ideological currents, political satire constitutes a distinct and essential category within Arabic meme culture. While at first glance it may not appear to represent a fixed ideological position, satirical memes occupy a unique space in political communication. As (Suryaningsih, 2025) notes, such memes often draw on irony, humor, and cultural references to critique power and expose contradictions across the political spectrum. Their apparent ambiguity allows them to resonate with diverse audiences, simultaneously challenging authority and mocking political extremes. Far from being ideologically neu-

Translation:
Metaverse means your mother removes her Hijab

Translation:
-Brekaing: Spokesperson for Maghreb convoy Omar shanoufy: We are close to the Libyan city of Misrata and we ask Egyptian authorities for safe passage to Rafah
-(Saracstically): Ok, Inch'allah

a)Political satire meme

b)Pan-Arabist meme

Translation:
-Arab country: We discovered oil!!
-USA:

Translation:
-If you want to apply Shariaa law just go to Afghanistan
-Afghanistan is merely a speck in a large sea tral, this flexibility enables political satire to serve as a powerful form of social commentary. For this reason, political satire warrants recognition as its own category in the analysis of Arabic political memes, rather than being placed under other ideological labels.

Figure 1: A sample of memes representing different political ideologies.

To address the aforementioned gap, we introduce ArPoMeme, the first dataset of approximately 7,300 Arabic political memes categorized by ideological orientation (Figure 1). Our work makes three primary contributions. First, we present a dataset structured around four categories:leftist, Islamist, pan-Arabist, and political satire and therefore providing a resource to study the ideological spectrum in Arabic digital discourse.

Second, we propose a novel classification method grounded in “self-identification,” which builds on the communal nature of meme production and circulation. Prior research emphasizes that memes resonate and spread only when their content aligns with the worldview of their originating community (Segev et al., 2015; Gal et al., 2016). This principle allows us to trust that memes emerging from explicitly ideological communities such as Facebook pages that label themselves as “leftist,” “pan-Arabist,” or “Islamist” which are representative of that ideological stance. Accordingly, instead of relying on machine learning models to infer ideological orientation from multimodal features, we directly attribute labels based on how meme pages self-identify through their bios, naming practices, or explicit positioning. This approach ensures that ideological categorization is grounded in the self-presentation of the communities producing and disseminating memes, rather than externally imposed classifications.

Third, we lay the groundwork for computational analysis of ideological narratives, framing strategies, and the circulation of political humor and critique in Arabic contexts. By anchoring classification in community self-identification, ArPoMeme opens the way for systematic studies of how distinct ideological currents leverage memes to frame issues, construct enemies and allies, and mobilize audiences.

In addition to the categorical labeling of ideological communities, we will also annotate the dataset to explore patterns of political polarization across these meme categories. This annotation will assess how memes visually and textually construct ideological antagonism, “us vs. them” framings, and calls to action, providing an empirical foundation for analyzing polarization in Arabic political humor and discourse. Such analysis will complement the dataset’s structural design by illustrating how ideological divisions manifest through multimodal narrative devices within Arabic meme culture.

By releasing this dataset, we aim to enable systematic analysis of how memes function as ideological texts and to provide a foundation for future research into political communication, digital culture, and computational social science in the Arab world.

## 2. Related Work

Research on memes as multimodal texts combining language and imagery has expanded rapidly across NLP and social computing. Within the Arabic context, this line of inquiry is still emerging, with recent efforts focusing mainly on harmful or manipulative dimensions of meme content. The first major contribution is ArMeme (Alam et al., 2024), a large-scale dataset of 129K Arabic memes annotated for propaganda detection across multiple platforms. The authors applied multi-stage filtering and OCR-based extraction, producing around 6K annotated memes that have since served as a benchmark for unimodal and multimodal classification tasks. Building on this, (Alam et al., 2025) introduced Propaganda to Hate, extending the ArAIEval-2024 corpus with fine-grained hate labels. Their multi-agent LLM annotation pipeline (GPT-4o, Gemini Pro, Claude 3.5 Sonnet) illustrated both the scalability and challenges of automated annotation in low-resource meme analysis. Within the same ArAIEval-2024 shared task, (Shah et al., 2024) proposed MemeMind, a system targeting multimodal propagandistic meme classification across three subtasks: text-only, image-only, and multimodal fusion. The authors fine-tuned state-of-the-art Arabic language models alongside vision architectures including ResNet18, EfficientFormerV2, and ConvNeXt-tiny, and leveraged generative augmentation via ChatGPT-4 and DALL-E-2 to address data scarcity. Their fusion of ConvNeXt-tiny and BERT in a late-fusion layer yielded notable gains in binary classification, underscoring both the promise and the remaining challenges of multimodal propaganda detection in Arabic.

Arabic NLP has also produced a growing body of resources targeting harmful and politically charged discourse more broadly. These include multi-label hate speech datasets (Zaghouani et al., 2024), multi-dialectal hate speech corpora (Charfi et al., 2024a), and stance detection resources (Charfi et al., 2024b), collectively establishing the annotation frameworks and taxonomies that inform our own labeling methodology. Studies of political framing on Arabic social media further motivate our work: Shestakov and Zaghouani (2024) examined digital framing around the Sheikh Jarrah evictions, Laabar and Zaghouani (2024) annotated stance, sentiment, and emotion in Facebook comments surrounding Tunisia's July 25 political measures. Together, these works demonstrate the richness and contentiousness of Arabic political discourse online and the utility of multi-dimensional annotation schemes for capturing it. Research on affective and emotional dimensions of Arabic social media (Shurafa and Zaghouani, 2024; Mohamed and Zaghouani, 2024) further highlights the role of sentiment and emotion in shaping political engagement, while earlier corpus efforts (Rangel et al., 2020) established foundational resources for demographic and dialectal analysis of Arabic social media language.

Beyond Arabic, several multilingual and cross-domain meme datasets have shaped the broader landscape of multimodal understanding. MultiOFF (Suryawanshi et al., 2020) provided one of the earliest multimodal resources for detecting offensive memes, combining textual and visual cues to address the subtleties of sarcasm and implicit hate. MemoSen (Hossain et al., 2022) introduced a multimodal sentiment analysis dataset in Bengali, highlighting the need for low-resource meme resources. Likewise, MET-Meme (Xu et al., 2022) incorporated metaphorical features into meme interpretation, demonstrating that metaphoricity enhances sentiment and semantic understanding. Complementarily, (Alzu'bi et al., 2023) proposed a multimodal deep learning approach with discriminant descriptors to detect offensive memes, achieving strong performance across metaphor, sentiment, and intention classification tasks. Together, these datasets emphasize how multimodal integration advances computational meme analysis. Finally, work such as (Widyaningsih et al., 2025) demonstrates how political memes during Indonesia's 2024 presidential election reinforced partisan identities through visual humor and symbolism. Inspired by such findings, ArPoMeme extends the study of polarization to Arabic meme ecosystems through dedicated multimodal annotation of antagonism and mobilization cues. Despite these advances, Arabic meme research remains limited to propaganda and hate detection. No existing resource systematically captures the ideological diversity that defines Arabic political discourse that spans leftist, Islamist, pan-Arabist, and satirical movements. ArPoMeme addresses this gap by introducing the first Arabic multimodal dataset focused on ideological orientation and polarization.

## 3. Data Collection and Processing

Our data collection process involves several steps as highlighted in Figure 2. The process followed a semi-automatic pipeline designed to target ideologically explicit meme sources. In contrast to large-scale scraping of heterogeneous content, we focused on Facebook pages that explicitly self-identified with a particular ideological leaning. This allowed us to ground our dataset in communities that explicitly position themselves within recognizable ideological currents. Specifically, we targeted pages aligned with four categories:

- Leftist pages (e.g., □·' l.LJ _□l_s', □ .,.lJl )
- Pan-Arabist pages (e.g., □·.,.. _-· I.... )
- Islamist pages (e.g., □·.,..>□,l I....)
- Political satire pages (e.g., □...,l l., □L.> )

| Page | Category | ID Element |
|---|---|---|
| Alhudood[1] | Political Satire | Bio |
| □·' l.LJ _□l_s' □_□.,.·.. "Infiltrated leftist cadres"[2] | Leftist | naming practice |
| □_.._· I.-_. "Nationalist memes"[3] | Pan-Arabist | naming practice |
| □·.,..>□l I.... "Islamic memes"[4] | Islamist | naming practice |

Table 1: Sample of the meme pages used in this dataset and its self identification elements

By drawing on communities that openly self-identify through their page names, bios, or content framing, we ensure that the memes collected authentically reflect the ideological stance of their originators.

Beyond qualitative source selection, we also quantified the dataset distribution across ideological categories to assess representational balance. As shown in Table 3, political satire pages account for the largest portion of the dataset, reflecting their prevalence and engagement levels across Arabic social media. Leftist pages represent the second-largest share, followed by Islamic and Pan-Arabist sources.

| Ideological Category | Number of Memes |
|---|---|
| Political Satire | 3,731 |
| Leftist | 2,608 |
| Islamist | 788 |
| Pan-Arabist | 235 |

Table 2: Distribution of memes by ideological category.

This distribution analysis helps situate our dataset within the broader landscape of Arabic online political expression.

### 3.1. Semi-Automatic Screenshot Collection

Facebook's restrictions on bulk image downloads prevented us from directly acquiring meme images through APIs or large-scale crawlers. To address this limitation, we developed a semi-automated screenshot-based collection pipeline using the Playwright browser automation framework, integrated with Google Drive for storage.

The pipeline functioned as follows:

1. Session Authentication: A secure login session was saved locally, enabling automated navigation across Facebook albums without repeated manual logins.
2. Bounding Box Selection: At the start of each run, a bounding box was manually drawn over the screen region containing meme content. This ensured that screenshots consistently captured only the relevant content area, excluding surrounding interface elements.
3. Automated Navigation and Capture: The script then navigated through the album using keyboard commands, capturing sequential screenshots of all memes displayed. Screenshots were automatically named using UUIDs to prevent duplication.
4. Cloud Storage Integration: Each screenshot was uploaded to a dedicated subfolder on Google Drive corresponding to the ideological source page. This ensured centralized, organized storage and facilitated subsequent annotation and processing.
5. Logging and Resume Functionality: A log was maintained for each album, recording the last captured URL. This enabled resumption of interrupted collection sessions without redundancy.

This hybrid manual–automated workflow offered both precision (via manual bounding box calibration) and scalability (via automated capture and upload), allowing us to build a balanced dataset across ideological categories.

### 3.2. Removing Non-Meme Images and Text Extraction

We extracted text from meme images using the Qwen2.5-VL-7B-Instruct vision-language model (Bai et al., 2025), a state-of-the-art multimodal model designed for visual understanding and text recognition tasks. This model was selected for its superior performance in detecting stylized text commonly found in memes, including text with various fonts, colors, overlays, and effects that traditional OCR systems often fail to recognize.

The extraction process involved loading images from all subfolders within the designated directory and processing them through the Qwen2.5-VL model with the prompt: "Extract and transcribe all the text you can see in this image." Each image was processed individually, and the model generated text outputs with a maximum token limit of 256. To prevent data loss during long processing sessions, results were automatically saved to a CSV file after every 10 images processed. The final dataset contained two columns: (1) filename - the full file path of each image, and (2) text - the extracted textual content. Images without detectable text were labeled as "NO TEXT." The process was designed to be resumable, allowing the script to skip previously processed images if interrupted, ensuring efficient handling of large-scale meme datasets.

We removed 3075 memes due to "NO TEXT" label. A manual verification of a sample 200 memes revealed that the model Utilized to extract text achieved a 83,2% accuracy.

## 4. Annotation Guidelines

This annotation task aims to measure political polarization in Arabic political memes through a mul-timodal lens that jointly considers textual and visual elements. Polarization is defined as communication that intensifies political division, fosters antagonism between groups, or encourages confrontational action. Recent literature highlights three interrelated dimensions that explain how polarization manifests in digital discourse. First, polarization often takes the form of "Us-vs-Them" framing, where political opponents are viewed as enemies rather than legitimate adversaries. As Hrbková et al. explain, many societies are transitioning from agonistic pluralism—where disagreement coexists with mutual legitimacy—to antagonistic struggle, where political out-groups are dehumanized. This dynamic has been documented in Arabic digital contexts, where polarization manifests across platforms through rhetorical strategies such as sarcasm, offensiveness, and oppositional stance-taking (Heraki and Zaghouani, 2025). Second, affective or emotional polarization deepens through resentment and hatred toward out-groups. Nettasinghe et al. demonstrate that such hostility disrupts compromise, leading to self-reinforcing cycles of division. Studies of Arabic social media further confirm that political flashpoints generate strongly polarized sentiment and hostile discourse (Laabar and Zaghouani, 2024). Finally, polarization can escalate into explicit calls to action, where shared identity and grievance motivate collective mobilization. Bliuc et al. find that intra-group reinforcement and radical narratives increase intentions for collective or confrontational behavior. Together, these dimensions (**Us-vs-Them Framing**, **Hostility Toward Out-Group**, and **Explicit Call to Action**) form the analytical foundation for identifying and measuring polarization in Arabic political memes.

[1] www.facebook.com/AlHudoodNet/
[2] www.facebook.com/Kimorganization/
[3] www.facebook.com/profile.php?id=100089592260983
[4] www.facebook.com/profile.php?id=100063604605426

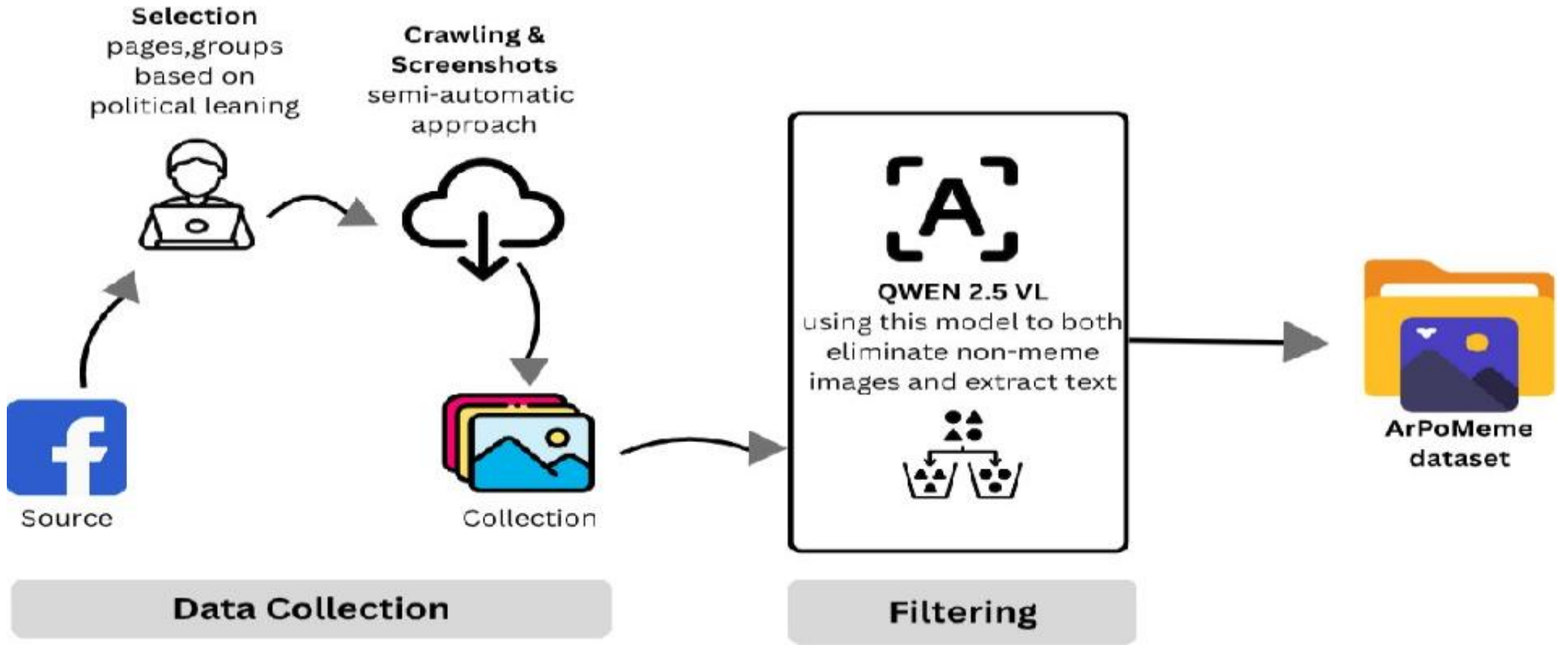

Figure 2: Data curation pipeline

### 4.1. Dimensions of Polarization (Binary, 0/1)

Annotators evaluate both text and image to determine whether each dimension is present. Each meme receives a binary label (0 = No, 1 = Yes) for the following three dimensions:

- **Us-vs-Them Framing:** The meme constructs two opposing groups or identities in conflict, either textually or visually.
- **Hostility Toward Out-Group:** The meme expresses insult, humiliation, or dehumanization toward a target group, through language or imagery.
- **Explicit Call to Action:** The meme urges collective action against a group (e.g., to resist, punish, or boycott), via direct language or symbolic imagery.

### 4.2. How to Apply Multimodal Annotation

Annotators must consider both textual and visual cues when labeling. Examples of multimodal indicators include:

- **Us-vs-Them Framing:** Text such as u· _b·; t"".," ("we are against them") or u· l.__Jl ,>L"" ("traitors to the nation"); visual cues such as opposing flags, crowds versus enemies, or split-group imagery.
- **Hostility Toward Out-Group:** Verbal insults like □·'__.>· ("traitors") or □·Jl.. ,..> ("scum"); mocking edits, crossed-out faces, animal depictions (e.g., pigs, rats), or violent imagery.

- **Explicit Call to Action:** Action verbs such as l__l·'l.∸ (“fight”) or l__-1.l.∸ (“boycott”); raised fists, protest symbols, or arrows suggesting mobilization.

### 4.3. Binary Annotation Rules

**Dimension 1: Us-vs-Them Framing (0/1)** Assign 1 if the meme divides people into opposing camps using identity framing. *Examples:*

- 0 – ""lå J l-. l ©®'l' (“prices are rising quickly”) (no group conflict)
- 1 – Text: ·'_b·J l t"".. _-L J l l..:.>l (“we are the people and they are the traitors”); Image: split layout with national flags or group separation

**Dimension 2: Hostility Toward Out-Group (0/1)** Assign 1 if clear hostility is expressed verbally or visually. *Examples:*

- 0 – l.k.> l.. LJl (“politics is wrong”) (normal criticism)
- 1 – Text: '.> l__.>· l (“the Brotherhood are traitors”) or u· ".'l..l.-Jl · l. b·; (“the filth of secularists”); Image: target group as animals or criminals

**Dimension 3: Explicit Call to Action (0/1)** Assign 1 if the meme urges direct or implied action through text or imagery. *Examples:*

- 0 – Opinion only
- 1 – Text: I""l.-Jl. 1>"." l l_.-1.l.· (“boycott the corrupt media”) or ,>..-Jl l..I1.l (“expel the traitors”); Image: raised fist with group symbol or protest arrow

### 4.4. Examples Across Ideological Categories

The following illustrate how the three dimensions appear across ideological orientations:

- **Leftist:** 0_.-.Jl.Ö.llJl Èl..-Jl 1 l_ – (“the capitalists sucked the workers’ blood”) with image of businessmen as vampires (Us-vs-Them = 1, Hostility = 1, Call to Action = 0)
- **Islamist:** Quran verse with ·-'.Iå·Jl H_l.g u·.. :.. l ...,"": __i- (“whoever fights the Sharia is an enemy of the nation”) (1, 1, 0)
- **Pan-Arabist:** Map and 0· __-·IåJ. ,>..-Jl ¼I·"· u·J l.. -' l. l...l (“we will not let the traitors steal our homelands”) (1, 1, 1)
- **Political Satire:** Politician edited as a clown (0, 1, 0)

### 4.5. Annotation Tool and Procedure

Annotators examined each meme as a **multimodal unit**, considering both its *textual* and *visual* elements together. They **identified the explicit or implied target group** and then **assigned binary values (0 or 1)** for each of the three annotation dimensions: *Us vs. Them framing*, *Hostility*, and *Call to Action*. In cases of uncertainty, annotators were instructed to **select 0 as the conservative option** to ensure labeling consistency. The label *Not Political* was applied **only when a meme clearly lacked political relevance**, ensuring that non-political humor or general cultural content was excluded from ideological analysis.

### 4.6. Collaborative Annotation Tool

To support the large-scale and multimodal annotation process, we developed a custom Streamlit-based web application tailored to the *ArPoMeme* dataset. The tool balances collaboration, efficiency, and reliability in labeling Arabic political memes across multiple annotators.

**Collaborative Design.** Annotators log in with unique identifiers, enabling personalized settings and synchronized progress. The system assigns distinct image batches to each user, allowing multiple annotators to work concurrently without overlap and ensuring dataset consistency.

**Batch-Based Workflow.** Each annotator receives a fixed number of memes (typically 50 per batch) drawn from the global pool. Batches are stored locally (outside shared drives) to allow offline work and avoid synchronization delays. Upon completion, annotations automatically sync to the main dataset and update the global progress tracker. This “local-first” approach maintains both speed and centralized integrity.

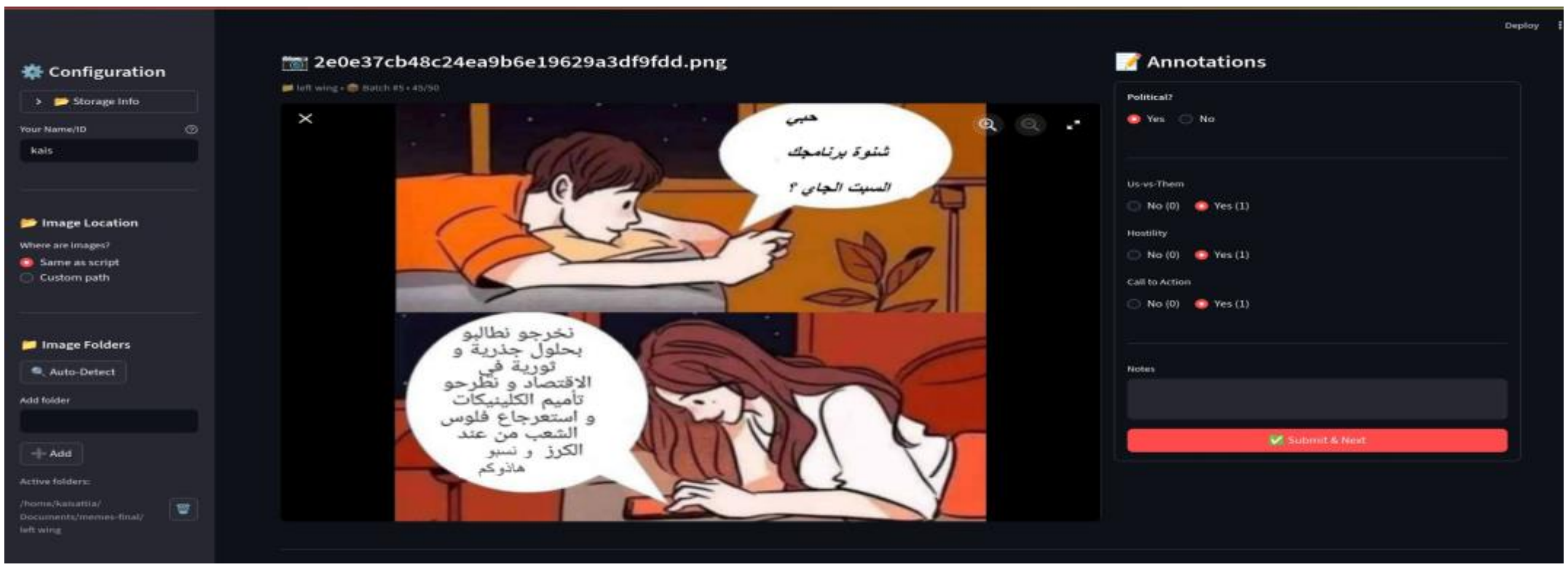

Figure 3: Custom Annotation Tool

**Progress Tracking and Version Control.** Every annotation is timestamped and linked to the annotator's ID, batch number, and image hash for traceability. Real-time metrics display batch completion rates, total images annotated, and active assignments. Local persistence ensures that incomplete sessions can be safely resumed after restarts or network interruptions.

**User Interface and Features.** The interface presents each meme alongside binary annotation fields for the three polarization dimensions: **Us-vs-Them Framing**, **Hostility Toward Out-Group**, and **Explicit Call to Action**. Annotators can also add notes, adjust settings for image paths, and review embedded guidelines via an integrated sidebar.

**Advantages.**

- **Collaborative:** Multiple annotators work in parallel on distinct batches.
- **Batch System:** Fixed-size subsets balance workload and aid quality control.
- **Local-First:** Data saved locally for speed, then synchronized to the repository.
- **Progress Monitoring:** Real-time tracking at user and project levels.
- **Auto-Sync and Recovery:** Automatic uploads and session resumption.

This tool (Figure 3) streamlined the annotation workflow, ensured transparent supervision, and enabled distributed participation in building the *ArPoMeme* dataset. The combination of local caching, batch versioning, and Google Drive synchronization improved both annotation efficiency and reliability. The accompanying annotation tool will be freely available with the dataset for academic research.

## 5. Data Annotation Quality Control

This section describes data annotation quality and measurement across manual annotation and model performance.

### 5.1. Inter-Annotator Agreement

To assess annotation reliability, we measured inter-annotator agreement (IAA) on a stratified sample of 100 memes (25 from each ideological category), independently annotated by three annotators. Fleiss' kappa was calculated for each polarization dimension and the political relevance label:

- **Us-vs-Them Framing:** $\kappa = 0.52$ (Moderate agreement)
- **Hostility Toward Out-Group:** $\kappa = 0.51$ (Moderate agreement)
- **Explicit Call to Action:** $\kappa = 0.39$ (Fair agreement)
- **Political Relevance:** $\kappa = 0.27$ (Fair agreement)

The moderate agreement on the first two dimensions suggests that identifying antagonistic framing and hostility is reasonably consistent across annotators, though contextual and multimodal complexities remain challenging. The fair agreement on explicit calls to action reflects the relative rarity and subjective nature of this label in the dataset. The fair yet relatively low kappa for political relevance is attributable to a prevalence effect: raw full-agreement reached 95%, but the near-unanimous assignment of memes as political (>95% of items) inflates expected agreement under chance, deflating kappa independently of genuine annotator disagreement.

### 5.2. Evaluation Metrics

We evaluated three vision-language models—LLaMA-3.2-VL, MBZAI/AIN, and Qwen3-VL-8B-Instruct—in a zero-shot classification setting. For each polarization dimension, we designed detailed instruction prompts describing the classification criteria. Each model processed the meme image together with the dimension-specific prompt to produce binary predictions (0 or 1), optionally accompanied by textual explanations. No model fine-tuning was performed; all systems relied solely on prompt-based instructions and their pre-trained multimodal representations.

| Model | Dim. | Acc | Prec | Rec | F1 |
|---|---|---|---|---|---|
| LLaMA-3.2 VL | us_vs_them | 0.553 | 0.553 | **1.000** | 0.713 |
| | hostility | 0.343 | 0.343 | **1.000** | 0.511 |
| | call_act | 0.034 | 0.034 | **1.000** | 0.066 |
| MBZAI AIN | us_vs_them | 0.579 | **0.833** | 0.292 | 0.432 |
| | hostility | 0.609 | 0.453 | 0.678 | 0.543 |
| | call_act | 0.916 | 0.000 | 0.000 | 0.000 |
| Qwen3 VL | us_vs_them | **0.703** | 0.761 | 0.771 | **0.766** |
| | hostility | **0.622** | **0.512** | 0.680 | **0.584** |
| | call_act | **0.936** | 0.000 | 0.000 | 0.000 |

Table 3: Performance comparison of vision-language models across three polarization-related dimensions.

Model predictions were evaluated against a manually annotated subset of 549 political memes from the ArPoMeme dataset. Performance was measured using standard classification metrics, including accuracy, precision, recall, and F1 score. Table 3 summarizes the comparative results across the three polarization dimensions: *us-vs-them framing*, *hostility toward out-groups*, and *calls to action*. Overall, performance varies substantially across both models and dimensions. Qwen3-VL-8B-Instruct achieved the strongest results for *us-vs-them* framing (F1 = 0.766) and showed moderate effectiveness in detecting *hostility* (F1 = 0.584). MBZAI/AIN produced mixed results, with moderate performance for hostility detection (F1 = 0.543) but lower performance for *us-vs-them* framing (F1 = 0.432). LLaMA-3.2-VL exhibited unstable behavior, characterized by high recall but low precision across dimensions, resulting in weaker overall F1 scores. Notably, all three models failed to reliably identify the *call-to-action* category, yielding near-zero F1 scores despite relatively high accuracy due to class imbalance. These findings highlight the challenges that current vision-language models face in detecting nuanced polarization cues in multimodal political meme content.

### 5.3. Error Analysis

Figure 4 presents confusion matrices for the three classification dimensions: *us-vs-them*, *hostility*, and *call-to-action*. To better understand model behavior, we examined the explanations generated by the vision–language model for misclassified samples.

For the *us-vs-them* dimension, most errors were false negatives where the model failed to detect implicit group polarization. Explanations frequently stated that the meme "does not show an explicit us-vs-them dynamic" or that it focuses on individual behavior rather than group divisions. This indicates a strong reliance on explicit linguistic cues, while many political memes convey polarization indirectly through ideological references, cultural symbols, or shared context.

In the *hostility* dimension, the model produced numerous false positives. The explanations suggest that negative or critical language was often interpreted as hostility, even when the meme's tone was satirical or descriptive. At the same time, hostile intent expressed through sarcasm, irony, or symbolic imagery was sometimes overlooked.

The *call-to-action* dimension shows a different pattern. Many false negatives were explained as memes that "express an opinion without calling for action," suggesting that the model primarily detects explicit imperative language while missing indirect mobilization cues.

## 6. Analysis

Figure 5 illustrates the distribution of polarization dimensions across ideological categories. Overall, the results indicate that polarization is most salient in *Islamic* memes, where **66%** exhibit clear *Us-vs-Them* framing and **39%** display *Hostility Toward an Out-Group*. In contrast, *Left Wing* and *Pan-Arabist* memes show moderate levels of antagonistic framing, with both scoring around **45%** on the *Us-vs-Them* dimension and lower hostility levels of **24%** and **33%**, respectively.

*Political Satire* occupies an intermediate position, with **57%** of memes engaging in oppositional framing and **29%** expressing hostility. This suggests that satire, while often humorous, remains an important vehicle for political critique and confrontation. Across all ideological categories, the presence of *Explicit Calls to Action* remains limited—ranging only from **3–5%**—indicating that most political Arab memes articulate polarization symbolically or rhetorically rather than mobilizing explicit collective action.

Figure 6 highlights the proportion of non-political memes identified within each ideological category. Annotation results reveal that such instances were relatively rare, with **3%** in Left Wing, **4%** in Political Satire, and **11%** in Pan-Arabist memes. Notably, none were found in the Islamic category. These findings underscore the effectiveness of the annotation process in distinguishing politically relevant content, while also reflecting the occasional presence of culturally themed or apolitical humor within ideologically aligned pages.

**Summary:** The results demonstrate measurable variation in polarization intensity across ideological currents. While antagonistic framing and hostility are prevalent features, explicit mobilization remains rare. The presence of non-political memes within certain categories highlights the nuanced boundaries between ideological expression and general cultural humor in Arabic meme ecosystems.

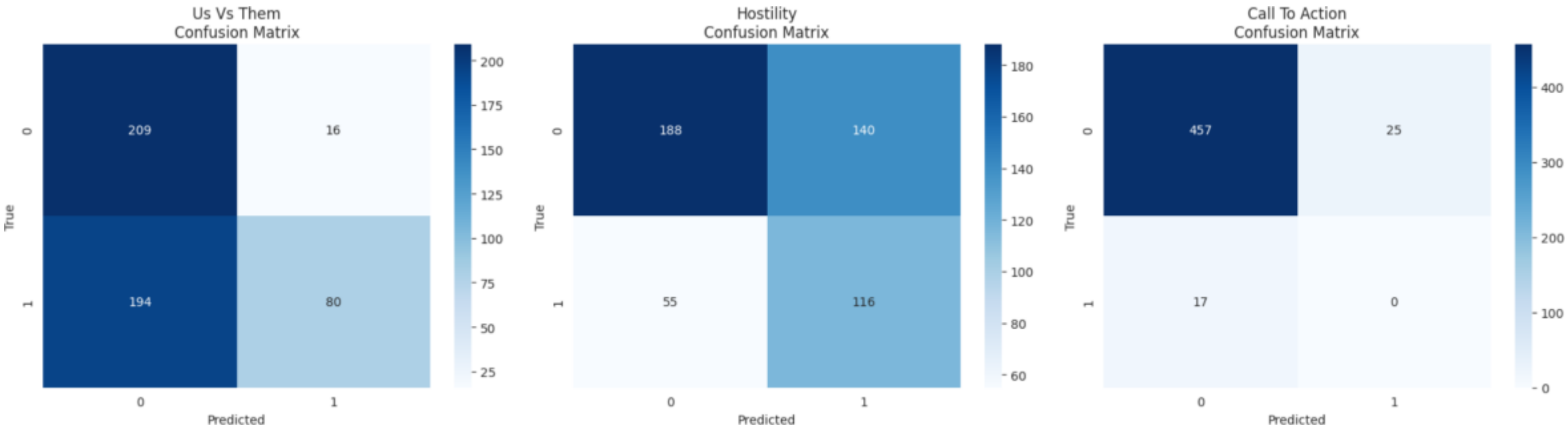

Figure 4: Error analysis for the Qwen3-VL model across Polarization dimensions.

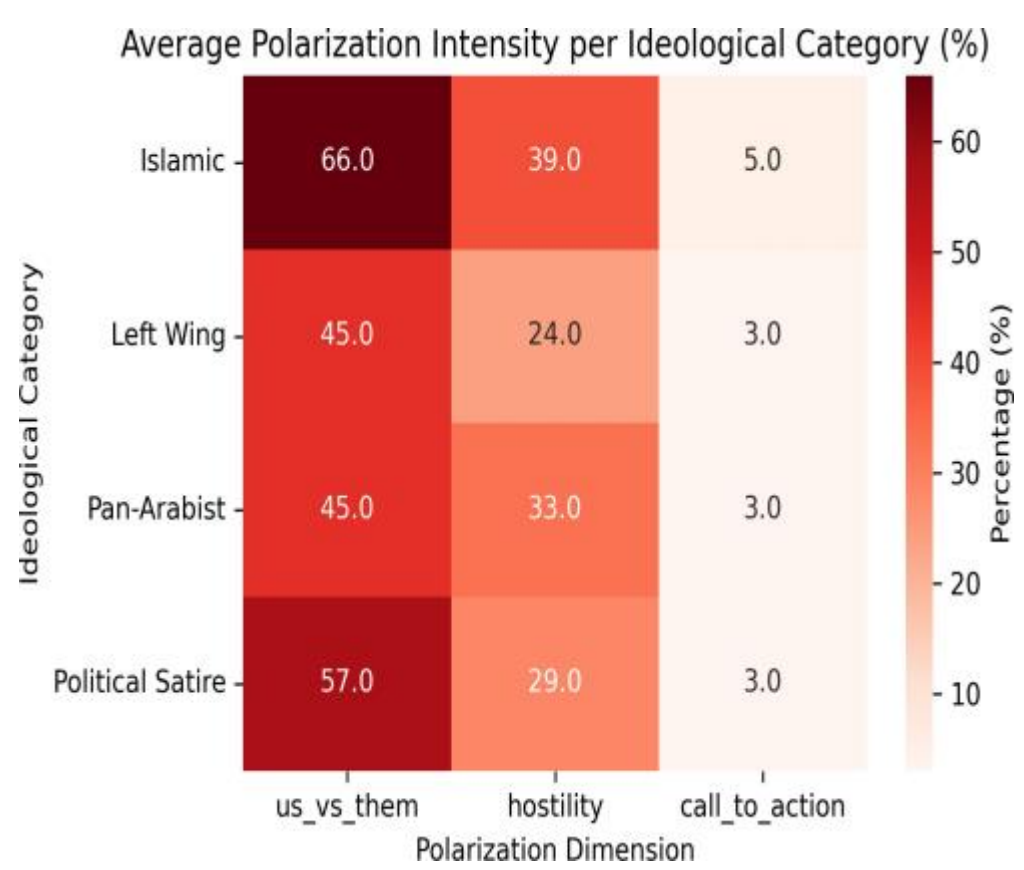

Figure 5: Polarization dimensions across ideological categories (%).

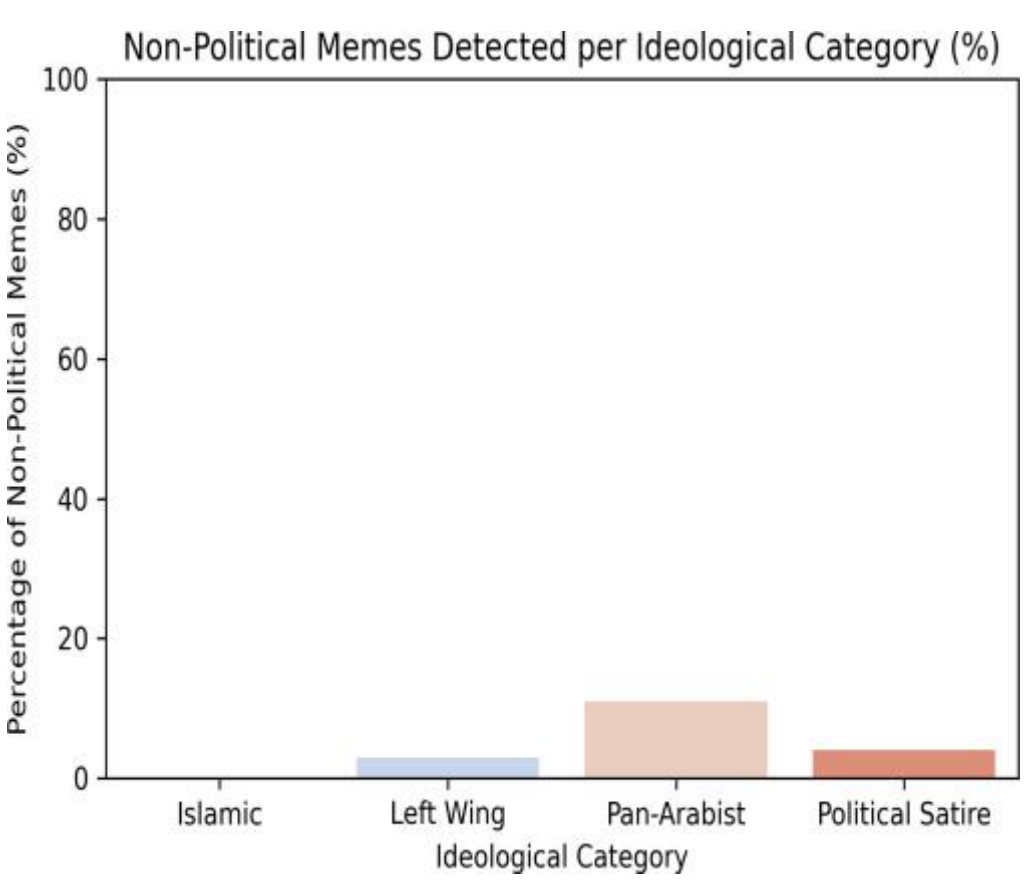

Figure 6: Share of non-political memes in each ideological category (%).

## 7. Data and Code Availability

The ArPoMeme dataset is publicly available. Access to the dataset requires completion of a request form[1]. The dataset is released strictly for research purposes. All materials are distributed under the *Creative Commons Attribution–NonCommercial–ShareAlike 4.0 International License (CC BY-NC-SA 4.0)*. Users must comply with the terms of this license, including proper attribution, non-commercial use, and distribution of derivative works under the same license.

## 8. Broader Impact

The *ArPoMeme* dataset offers an important resource for studying how visual and textual elements interact to shape political communication in the Arabic-speaking world. By providing a transparent and reproducible multimodal dataset, this work supports research on online discourse, ideological framing, and polarization across different cultural contexts. Beyond computational linguistics, the dataset can also inform studies in media analysis, visual communication, and digital humanities, fostering interdisciplinary collaborations between language technology and the social sciences. Through its open release, *ArPoMeme* contributes to a more inclusive research landscape in Arabic NLP and encourages comparative studies on political expression across languages and regions.

## 9. Conclusion

This paper introduced *ArPoMeme*, a large-scale dataset of Arabic political memes designed to advance the study of ideological polarization in multimodal discourse. By combining automated collection, human verification, and structured annotation, the dataset provides a comprehensive resource for analyzing the intersections of language, imagery, and ideology. Preliminary findings reveal distinct polarization patterns across ideological categories,

[1] https://forms.gle/W7xpLt7io326bR3A6

highlighting the communicative strategies underlying antagonistic framing and mobilization.

*ArPoMeme* contributes to the development of tools and methodologies for multimodal analysis in Arabic, fostering collaboration across natural language processing, digital humanities, and computational social science. The dataset, code, and annotation guidelines will be released for research use, supporting reproducible studies on digital culture and political discourse in Arabic-speaking communities.

## 10. Limitations

A key challenge in the construction of ArPoMeme lies in distinguishing memes from non-meme images within the collected data. To address this, we employed an automatic filtering step based on text detection: images that did not contain detectable overlaid text were classified as non-memes and excluded. While this method provides a scalable solution in the absence of extensive manual annotation, it is not without drawbacks.

Certain non-meme images may nonetheless contain text that was pre-integrated into the original content (e.g., screenshots of articles, advertisements, or infographics). Such content can be misclassified as memes, despite not being the product of meme-making practices or fitting the cultural conventions of meme discourse.

Due to limited annotation resources, it was not feasible to manually review and remove all such cases across the dataset. As a result, ArPoMeme may still contain a subset of non-meme images. While we anticipate that this proportion remains relatively small, it is important for researchers utilizing the dataset to be aware of this limitation. Future work could refine meme detection through more sophisticated multimodal models that combine text detection with structural and stylistic features specific to memes.

Another limitation concerns the uneven distribution of memes across ideological categories, as discussed in Section 3. Political satire and leftist memes are heavily represented compared to pan-Arabist and Islamist ones. This imbalance may not only reflect real disparities in meme production or engagement levels across Arabic online communities but could also result from algorithmic bias in Facebook's content recommendation and visibility mechanisms. During manual selection, our researchers were exposed to pages surfaced by Facebook's algorithmic ranking, which tends to prioritize highly engaging and widely shared content. Consequently, progressive and politically satirical pages that are often characterized by high interaction rates and shareable humor,were more prominently displayed, while ideologically conservative or pan-Arabist meme pages may have remained underrepresented due to lower algorithmic visibility or stricter moderation environments.

This bias underscores the structural challenges of collecting politically balanced datasets from social media platforms that mediate visibility through engagement-driven algorithms. Future iterations of ArPoMeme could mitigate this limitation by incorporating alternative data collection strategies, such as direct keyword-based search, cross-platform sampling, or curated lists of lesser-known pages to achieve a more even ideological representation.

A further challenge arises from the inherently multimodal and context-dependent nature of memes, which complicates the annotation of polarization dimensions. Political meaning is often conveyed not only through textual content but also through visual symbolism, compositional cues, or layers of irony and sarcasm that resist literal interpretation. For instance, antagonism or hostility may be communicated visually (through edited faces, national symbols, or exaggerated caricatures) without explicit textual indicators. Similarly, satirical memes frequently employ humor and contradiction to critique both sides of a political divide, making it difficult to assign binary labels such as Us-vs-Them or Hostility Toward Out-Group without deeper contextual understanding.

These complexities highlight the limits of applying rigid binary annotation schemes to culturally and semiotically rich multimodal artifacts. Future versions of the ArPoMeme annotation framework should therefore explore more nuanced, hierarchical, and context-sensitive guidelines that better capture irony, visual rhetoric, and implied antagonism. Incorporating annotator training on multimodal interpretation, expanded label taxonomies (e.g., degrees of hostility or implicit vs. explicit framing), and iterative consensus-building methods could significantly enhance the reliability and interpretive depth of polarization analysis in Arabic meme research

## 11. Ethical Considerations

All content included in the *ArPoMeme* dataset was collected exclusively from **publicly accessible Facebook pages** that explicitly identify themselves as political, activist, or satirical sources. **No private data, user comments, or personal identifiers** were accessed or stored at any point in the collection process. Each meme was captured through automated screenshotting and stored in a secure environment with anonymized metadata. The collection and annotation procedures comply with **Meta's data access and privacy policies** and follow the principles of **responsible research and transparency**.

**Annotator well-being** was prioritized throughout the project. The annotation team consisted of native Arabic speakers who were trained on clear labeling guidelines and provided with regular check-ins to reduce fatigue and mitigate exposure to potentially distressing political imagery or language. Annotators had the option to skip any content perceived as offensive or emotionally difficult without penalty.

The project maintained strict **data minimization practices**: only the image content, page-level identifiers, and derived textual transcriptions were stored. All links to original posts were removed in the final release to prevent re-identification of individuals or small groups. Any offensive, hateful, or demeaning material was retained solely for research transparency and is clearly flagged in the dataset documentation.

The dataset and accompanying materials will be released under a **Creative Commons Attribution–NonCommercial–ShareAlike 4.0 license**, ensuring that they can be used for academic and non-commercial purposes while requiring attribution and preservation of the same open conditions for derivative works. The ethical design of *ArPoMeme* emphasizes **respect for privacy**, **responsible reuse**, and **transparency** to enable safe and reproducible research on Arabic political discourse without amplifying harmful narratives or bias.

## Acknowledgments

This work was made possible by the National Priorities Research Program (NPRP) grant NPRP14C-0916-210015 from the Qatar National Research Fund (QNRF), a member of the Qatar Research, Development and Innovation Council (QRDI).

## 12. Bibliographical References